\documentclass{article} 
\usepackage{iclr2022_conference,times}

\usepackage{amsmath,amsfonts,bm}

\def\eqref#1{equation~\ref{#1}}

\def\1{\bm{1}}

\DeclareMathAlphabet{\mathsfit}{\encodingdefault}{\sfdefault}{m}{sl}
\SetMathAlphabet{\mathsfit}{bold}{\encodingdefault}{\sfdefault}{bx}{n}

\usepackage{hyperref}
\usepackage{url}
\usepackage[ruled,vlined]{algorithm2e}
\usepackage{graphicx}
\usepackage{subcaption}
\usepackage{dsfont}

\title{Learning from demonstrations with SACR2: Soft Actor-Critic with Reward Relabeling}

\author{
Jesus Bujalance Martin \\
MINES ParisTech, PSL University, Center for robotics \\
60 Bd. Saint Michel 75006 Paris, France \\
\texttt{name.surname1\_surname2@mines-paristech.fr}
\And
Raphael Chekroun \\
MINES ParisTech, Valeo DAR \\
\texttt{name.surname@mines-paristech.fr} \\
\texttt{name.surname@valeo.com}
\And
Fabien Moutarde \\
MINES ParisTech, PSL University, Center for robotics \\
\texttt{name.surname@mines-paristech.fr}
}

\iclrfinalcopy 
\begin{document}

\maketitle

\begin{abstract}
During recent years, deep reinforcement learning (DRL) has made successful incursions into complex decision-making applications such as robotics, autonomous driving or video games. Off-policy algorithms tend to be more sample-efficient than their on-policy counterparts, and can additionally benefit from any off-policy data stored in the replay buffer. Expert demonstrations are a popular source for such data: the agent is exposed to successful states and actions early on, which can accelerate the learning process and improve performance. In the past, multiple ideas have been proposed to make good use of the demonstrations in the buffer, such as pretraining on demonstrations only or minimizing additional cost functions.
We carry on a study to evaluate several of these ideas in isolation, to see which of them have the most significant impact. We also present a new method for sparse-reward tasks, based on a reward bonus given to demonstrations and successful episodes. First, we give a reward bonus to the transitions coming from demonstrations to encourage the agent to match the demonstrated behaviour. Then, upon collecting a successful episode, we relabel its transitions with the same bonus before adding them to the replay buffer, encouraging the agent to also match its previous successes.
The base algorithm for our experiments is the popular Soft Actor-Critic (SAC), a state-of-the-art off-policy algorithm for continuous action spaces.
Our experiments focus on manipulation robotics, specifically on a 3D reaching task for a robotic arm in simulation. We show that our method SACR2 based on reward relabeling improves the performance on this task, even in the absence of demonstrations.
\end{abstract}

\section{Introduction}
\label{introduction}

Despite having known great success, reinforcement learning algorithms have yet to prove that they can consistently produce good results across different domains. Works like \cite{Engstrom2020Implementation} show that we still don't know with certainty what are the things that matter the most in current algorithms.
Other challenges such as reproducibility issues have brought some scepticism to the field.

Two of the main challenges in reinforcement learning are reward shaping and sample efficiency. Reward shaping makes it very difficult to translate results from one task to another, and often relies on the intuition of the designer rather than a robust methodology. 
Sample efficient algorithms are required to obtain faster and more reliable results, particularly in robotics where it is much harder to deploy an algorithm outside of simulation. Recent progress has enabled some deployment to real robots, but there are still issues to consider such as safety or human intervention \citep{DBLP:journals/corr/abs-2104-11203}.
The recent trend towards more data-driven algorithms could be a solution to both of these problems. Indeed, additional data can alleviate the need for online data collection, and demonstration data can guide the agent to good behaviours with a simple task-agnostic reward function.

In this work, we focus on a reaching task with sparse rewards, and aim to identify what are the best methods that leverage expert demonstrations to improve sample efficiency.
We focus on generic task-agnostic methods that can be applied to any off-policy algorithm with some minor modifications.
We also present a new such method, which is based on the observation that, in hindsight, a successful episode of collected experience is in fact a demonstration, so it should receive the same treatment. In particular, we propose to add a reward bonus to all transitions coming from both demonstrations and succesful episodes.
We instantiate our approach with Soft Actor-Critic (SAC) \citep{haarnoja2018soft}, and compare it to two other algorithms, SACfD \citep{DBLP:journals/corr/VecerikHSWPPHRL17} and SACBC \citep{nair2018overcoming}, presented in detail in section \ref{related_work}.

The contributions of this paper are the following:
\begin{itemize}
    \item We present an ablation study of existing methods that improve sample efficiency  by leveraging demonstrations in off-policy reinforcement learning algorithms.
    \item We introduce a new such method that consists on giving a reward bonus to demonstrations and relabeling successful episodes as demonstrations. Our approach is among the methods with the highest performance increase, and is a component of the final algorithm that achieves the best results. It also consistently solves the task without demonstrations, whereas SAC often fails.
\end{itemize}

\section{Related work}
\label{related_work}

Reinforcement Learning (RL) and Imitation Learning (IL) are the two most popular paradigms to solve decision-making tasks within machine learning. In classical RL, the data is collected during training by interacting with the environment, which provides a reward signal that we try to maximize. In IL, the data is generally collected before training, and it consists of expert trajectories of a behaviour that we wish to copy or imitate.

\textbf{Learning for robotics.}

In this paper, we focus on general-purpose RL algorithms that can be applied to any problem out of the box. However, roboticists have for a long time pursued more sample-efficient algorithms that could be deployed in the real world.
Outside of RL, Transporters Networks \citep{zeng2020transporter} exploit spatial symmetries and RGB-D sensors to solve a variety of tasks from just a few samples, and NTP \citep{DBLP:journals/corr/abs-1710-01813} and the more recent NTG \citep{huang2019neural} use clever representations to execute a hierarchy of movement primitives that solve complex sequential tasks from demonstrations. The recent C2F-ARM from \cite{james2021coarse} relies on RL, point cloud inputs, and a clever pre-processing pipeline to achieve impressive results from just a handful of demonstrations. Many successful algorithms for robotics have come from model-based RL, since they can learn from task-agnostic data. Earlier works proposed dynamics models over latent representation spaces, which could be learnt from a reconstruction objective \citep{finn2016deep}, or directly from their ability to produce models that accurately explain the observed data \citep{zhang2019solar}. More recently, works such as \cite{finn2017deep} or \cite{schmeckpeper2020learning} successfully learnt a model directly in image space. More generally, the recent trend of offline RL has brought algorithms able to leverage huge amounts of data, expert or not, such as QT-Opt \citep{kalashnikov2018qt} or the most recent multi-task version MT-Opt \citep{kalashnikov2021mt}.

\textbf{Relabeling past experience.}

Since off-policy RL algorithms can theoretically use data coming from any policy, a natural idea was to share data between tasks in a multi-task setting. An even better idea came in Hindsight Experience Replay (HER) \citep{NIPS2017_453fadbd}, where the authors pointed out that if we accidentally solve one task when trying to perform another task, that experience is still optimal if we relabel the goal that was initially intended. Similar and more general works followed, such as GCSL \citep{ghosh2019learning}, Generalized Hindsight \citep{li2020generalized}, and HIPI \citep{eysenbach2020rewriting}, which reframes the relabeling problem as inverse RL. RCP \citep{DBLP:journals/corr/abs-1912-13465} extended the idea to the single-task setting, by learning a policy conditioned on the trajectory return: trajectories collected from sub-optimal policies can be viewed as optimal supervision for matching the reward of the given trajectory.

\textbf{Learning from demonstrations.}

The most straight-forward variant of IL is behaviour cloning (BC), where we directly look for a policy that acts like the expert by solving a supervised learning problem. This approach suffers from a series of issues, namely compounding errors that can put the agent into states where it can no longer recover, because they lie outside the training state distribution. However, BC has seen success in a variety of complex applications such as autonomous driving \citep{bojarski2016end}. 
More robust algorithms have been proposed, like Dagger from \cite{ross2011reduction}, which does however require that the expert is available during training.
Another variant of IL which has known great success is inverse RL (IRL), where we try to infer the reward function that the expert was most likely trying to maximize, while optionally jointly learning a policy. IRL algorithms benefit from online data collection, but assume that the reward signal from the environment is unknown. The most recent IRL algorithms that can handle complex behaviours are based on adversarial optimization, such as GCL from \cite{finn2016guided}, which presents a similar idea to the well-known GAIL from \cite{ho2016generative}.

\textbf{Learning from both demonstrations and reinforcement learning.}

Demonstrations can be used to design the reward, guide exploration, augment
the training data, initialize policies, etc. In NAC \citep{gao2018reinforcement}, the demonstrations, which can be sub-optimal, are used as the only training data during the first iterations. In \cite{zhu2018reinforcement} the demonstrations are used to augment the manually designed task reward with an imitation-based reward. In DAPG \citep{DBLP:journals/corr/abs-1709-10087} the demonstrations are used twice: to pretrain with behavior cloning, and to augment the policy gradient equation. 
In DAC \citep{liu2020demonstration}, they introduce a novel objective based on an augmented reward, the larger the closer the policy to the expert policy.

We will focus on three algorithms that can be applied to any continous-action off-policy algorithm with very minor modifications.

SACfD \citep{DBLP:journals/corr/VecerikHSWPPHRL17} (originally DDPGfD based on DDPG) introduces three main ideas: transitions from demonstrations are added to the replay buffer, prioritized replay is used for sampling transitions (demonstration data is given a bonus to be sampled more often), and a mix of 1-step and n-step return losses are used.

SACBC \citep{nair2018overcoming} (has no name and originally based on DDPG) also introduces three main ideas: transitions from demonstrations are added to a separate additional replay buffer, an auxiliary behaviour cloning loss is applied to samples from this buffer,
and some episodes are reset to a state sampled uniformly from a demonstration.
The authors additionally present other ideas regarding the multi-goal setting which we won't cover in this work.

SQIL \citep{reddy2020sqil} is actually a pure IL algorithm, since the reward signal is supposed unknown, but we present it here since it also incorporates demonstration data into the replay buffer. The replay buffer is initially filled with demonstrations where the rewards are always $r = 1$, and new experiences collected by the agent are added with reward $r = 0$. 

\section{Background}
\label{background}

\textbf{Reinforcement learning.}

In reinforcement learning, an agent interacts with an environment by performing an action and observing a feedback signal (reward $r$) and the new state of the environment. 
The goal is to find the policy (function mapping states $s$ to actions $a$) that maximizes the discounted cumulative reward:
\begin{equation}
    \pi^* = \arg\max\limits_\pi \mathbb{E}_{\tau \sim p_\pi(\tau)} \left[ \sum\limits_{k=0}^{T} \gamma^k r(s_{1+k},a_{1+k}) \right] 
    \label{objective}
\end{equation}
We define the \textit{Q-function} $Q^\pi(s_t,a_t) = \sum\limits_{t'=t}^{T} \mathbb{E}_{p_\pi} \left[ \gamma^{t'-t} r(s_{t'},a_{t'})|s_t,a_t \right]$ as the reward-to-go from the state $s_t$ if we pick the action $a_t$ and then follow $\pi$.

\textbf{Soft Actor-Critic.}

In reinforcement learning, the optimal policy is always deterministic under full observability, but stochastic policies have interesting properties: better exploration and robustness (due to wider coverage of states), and multi-modality.
We need an objective that promotes stochasticity by maximizing the entropy $\mathcal{H}$ of the policy (we omit $\gamma$ here for simplicity):
\begin{equation}
    \pi^* =  \arg\max_\pi \sum\limits_{t=1}^{T} \mathbb{E}_{(s_t,a_t) \sim p_\pi} \left[ r(s_t,a_t) + \alpha \mathcal{H}(\pi(\cdot|s_t))\right] 
\end{equation}
We can define a new Q-function (slightly different to accommodate the entropy term) which follows the soft Bellman equation:
\begin{equation}
    Q^\pi(s,a) = r + \gamma\mathbb{E}_{s' \sim p(\cdot|s,a),a' \sim \pi(\cdot|s')} \left[ Q^\pi(s',a') - \alpha\log\pi(a'|s')\right]
\end{equation}
To train the critic, we can approximate the right-hand expectation with samples, set it equal to $y$, and minimize the MSBE loss on a parameterized $Q_\phi$ (in practice, two approximators $Q_{\phi_1}$ and $Q_{\phi_2}$ are trained):
\begin{equation}
    \mathcal{L}_1(Q_\phi) = \underset{(s,a,r,s',d) \sim {\mathcal D}}{{\mathbb E}}\left[
    ( Q_{\phi}(s,a) - y )^2
    \right]
\label{sac}
\end{equation}
To train the actor $\pi_\theta$, the actor loss is derived from the reparameterization trick to compute samples $\Tilde{a}_\theta(s,\epsilon) = \mu_\theta(s) + \sigma_\theta(s)\epsilon$, where $\epsilon$ is some random noise:
\begin{equation}
    \mathcal{L}(\pi_\theta) = -\underset{s \sim {\mathcal D}, \epsilon\sim{\mathcal N}}{{\mathbb E}}\left[
    \min_{j=1,2} Q_{\phi_j}(s,\Tilde{a}_\theta(s,\epsilon))
    - \alpha\log\pi_\theta(\Tilde{a}_\theta(s,\epsilon)|s)
    \right]
\label{sac-actor}
\end{equation}

\section{Method}
\label{method}

We propose SACR2, Soft Actor-Critic with Reward Relabeling, a straight-forward method that can be implemented to any off-policy reinforcement learning algorithm with sparse rewards.
First, we add demonstration data to the buffer. The last transition of each expert trajectory is given the sparse reward from the environment. The other transitions are given a reward equal to $b$, typically smaller than the sparse reward.
Then, every time we collect a successful episode, we relabel the last $N-1$ transitions leading to the sparse reward, where $N$ is the average length of the demonstrations, by assigning to them a reward equal to $b$.

The first part of our algorithm is most similar to SQIL. 
Intuitively, it gives the agent an incentive to imitate the expert. Their paper shows theoretical connections between SQIL and regularized behaviour cloning.
One important difference is that SQIL is a pure imitation learning algorithm, while our method learns from both the reward bonuses and the reward from the environment. This also means that our reward bonus $b$ should be carefully tuned so that it has an impact without completely swallowing the environment reward. Empirically we found that $b \approx \frac{R}{N-1}$, where $R$ is the value of the sparse reward, provided good results (see section \ref{results}), meaning that an expert demonstration has on average a return twice as big as a successful episode.
Intuitively, SQIL also gives an incentive to avoid states that weren't in the demonstration data, which could potentially be harmful if those states led to successful behaviour.

The relabeling part of our algorithm tries to mitigate this issue and is most similar to Self-Imitation Learning (SIL) from \cite{Oh2018SIL}. In SIL, the self-imitation is achieved by an additional loss function that pushes the agent to imitate its own decisions in the past only when they resulted in larger returns than expected. In our method, the self-imitation is achieved by effectively treating successful episodes as if they were demonstrations. In order to avoid rewarding poor trajectories that solve the task by chance, we only reward the last $N-1$ transitions leading to the sparse reward, where $N$ is the average length of the demonstrations.

\begin{algorithm}
\SetAlgoLined
Require: $b$ reward bonus, $N$ average length of demonstrations\;
 Initialize buffer with demonstrations, set reward $r=b$ for all non-final transitions\;
 Initialize empty episode\;
 \While{not converged}{
  do a SAC update\;
 
  \If{len(episode) == 0}{
   collect one episode\;
     \If{episode is successful}{
   set $r = b$ for the last $N-1$ non-final transitions\;
   }
   }
   pop a transition $(s,a,s',r)$ from the episode and add it to the buffer\;
 }
 \caption{SACR2: Soft Actor-Critic with Reward Relabeling}
\end{algorithm}

\section{Experimental setup}
\label{experimental_setup}

\begin{figure}[ht]
\centering
  \begin{subfigure}[b]{0.495\textwidth}
    \includegraphics[width=\textwidth]{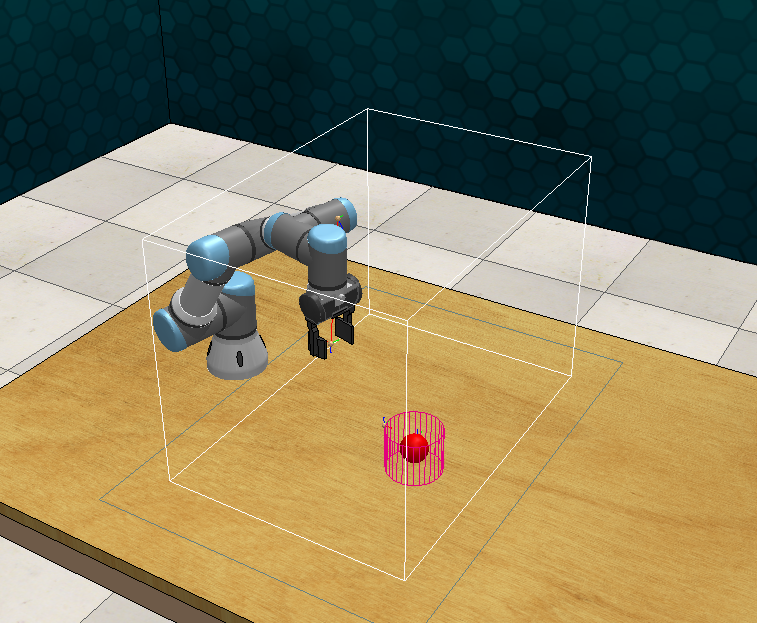}
  \end{subfigure}
  \begin{subfigure}[b]{0.47\textwidth}
    \includegraphics[width=\textwidth]{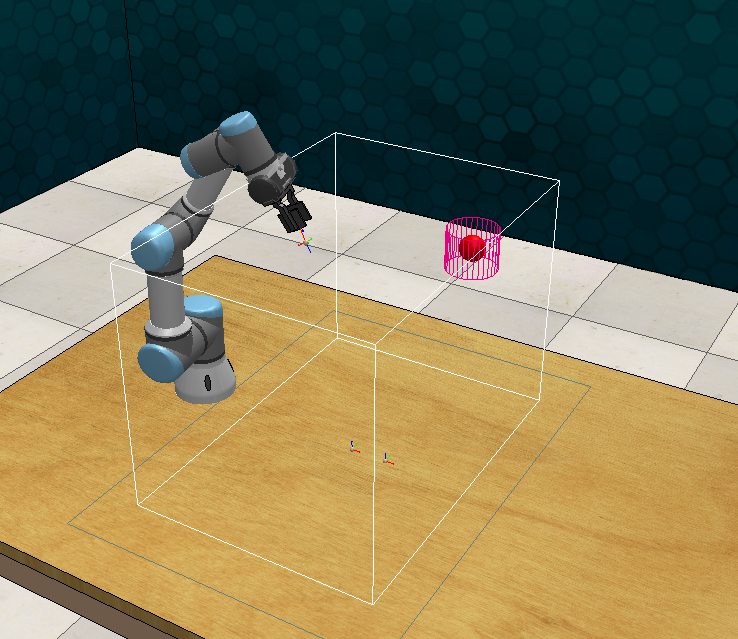}
  \end{subfigure}
  \caption{Two snapshots from two different episodes of the reaching task on RLBench. The target ball appears at the beginning of a new episode somewhere within the delimited box.}
  \label{robot}
\end{figure}

We evaluate our method in a simulated reaching task for a 6 degrees-of-freedom robot manipulator. 
The goal is to reach a target ball that appears randomly in the 3D space within the reach of the robot. The initial position of the target changes after each episode, but it doesn't move during an episode.
The initial state of the robot is always the same, close to an upright position.
An episode ends once the robot has reached the ball, or after 100 time-steps. 
The reward is fully sparse, and is equal to $+100$ if the robot reaches the ball and 0 otherwise. 
The state has 22 dimensions, 19 from the robot proprioceptive state (joint angles, joint speeds, gripper pose) and 3 from the task-related information (3D coordinates of the ball).

\textbf{Simulator and demonstration data collection}.

The reaching task is part of the benchmark and learning environment RLBench from \cite{james2020rlbench}, which is built around CoppeliaSim \citep{rohmer2013v}. The backend physics engine is the Bullet physics library \citep{coumans2015bullet}, and the expert demonstrations are provided by RLBench and rely on OMPL \citep{sucan2012open} for motion planning. 

\textbf{Experiments}.

The basic SAC algorithm without demonstrations is able to solve this task (with close to 100$\%$ accuracy) on some runs, so our goal is to reduce the amount of training steps required to get there. Each experiment is averaged over 4 runs.
We want to answer four questions:
Which of the ideas presented in SACBC and SACfD have the most significant impact ? How does SACR2 perform ? What is the configuration that achieves the best performance ? Does SACR2 make a difference when no demonstrations are available ?

In order to answer the first two questions, we test each method in isolation on top of a baseline SAC+Demo, which we define as SAC with demonstrations in the buffer. We apply the following modifications to both SAC+Demo and SACR2:
\begin{itemize}
\item Single buffer initially filled with 200 demonstrations, and kept thereafter at a ratio of $10\%$ demonstration data. After 130000 training iterations, around 800 more demonstrations have been added to the buffer.
\item We additionally add 1000 random interactions to the buffer before training, and we pre-train during 3000 iterations before collecting any data.
\item The data is sampled from the buffer according to prioritized experience replay (PER) \citep{DBLP:journals/corr/SchaulQAS15}.
\item The replay ratio on the collected data is set to 32. Since the batch size is set to 64, the agent takes two environment steps per training step.
\item As suggested in both SACfD and SACBC, we use L2 regularization losses on the weights of the critic and the actor.
\end{itemize}

We evaluate SACR2, four loss functions, two buffer configurations, pre-training on demonstrations, resetting some episodes to a demonstration state, and modifying the sampling probabilities of the replay buffer.

\section{Results}
\label{results}

\begin{figure}[ht]
\centering
  \begin{subfigure}[b]{0.495\textwidth}
    \includegraphics[width=\textwidth]{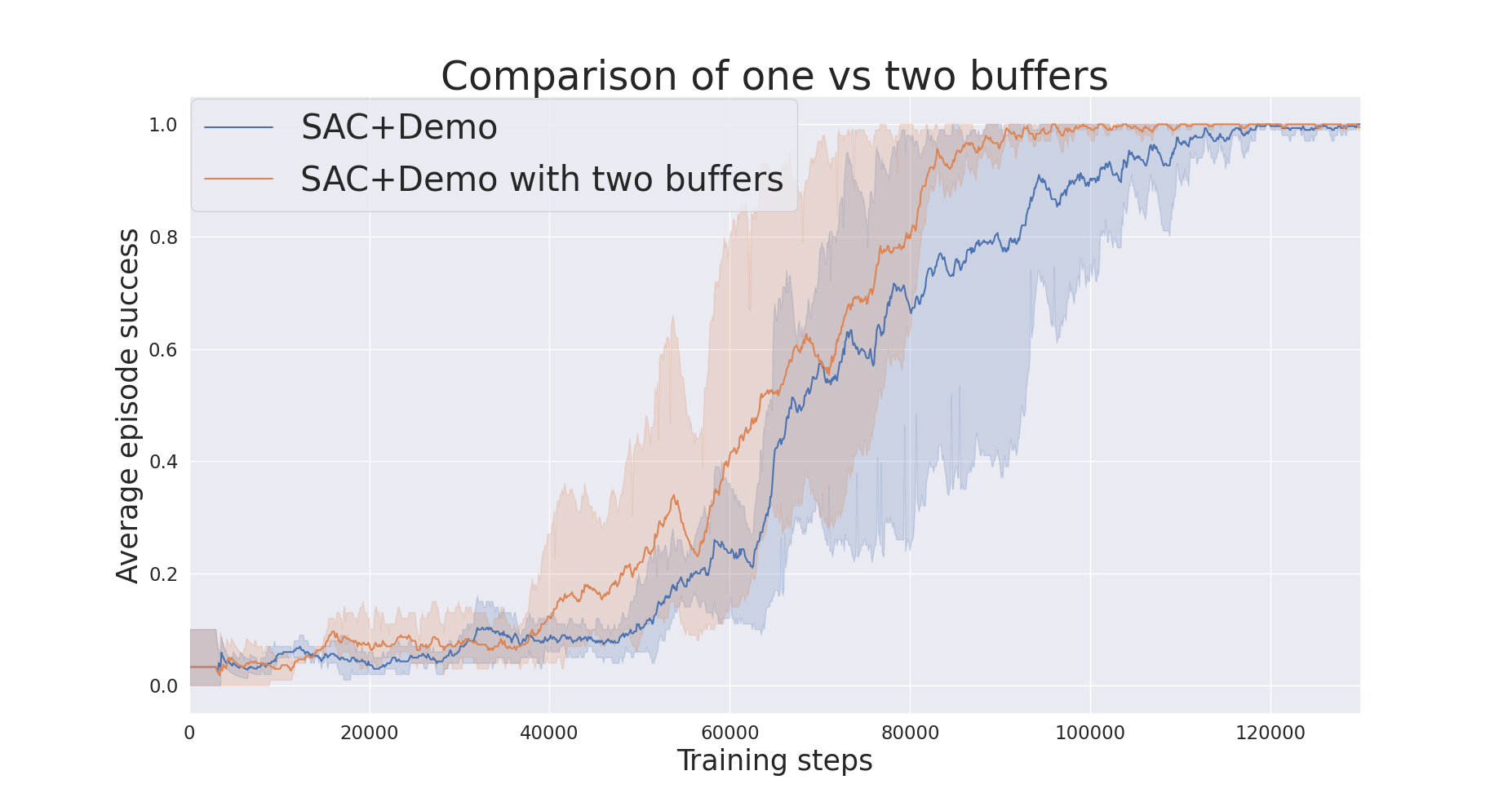}
  \end{subfigure}
  \begin{subfigure}[b]{0.495\textwidth}
    \includegraphics[width=\textwidth]{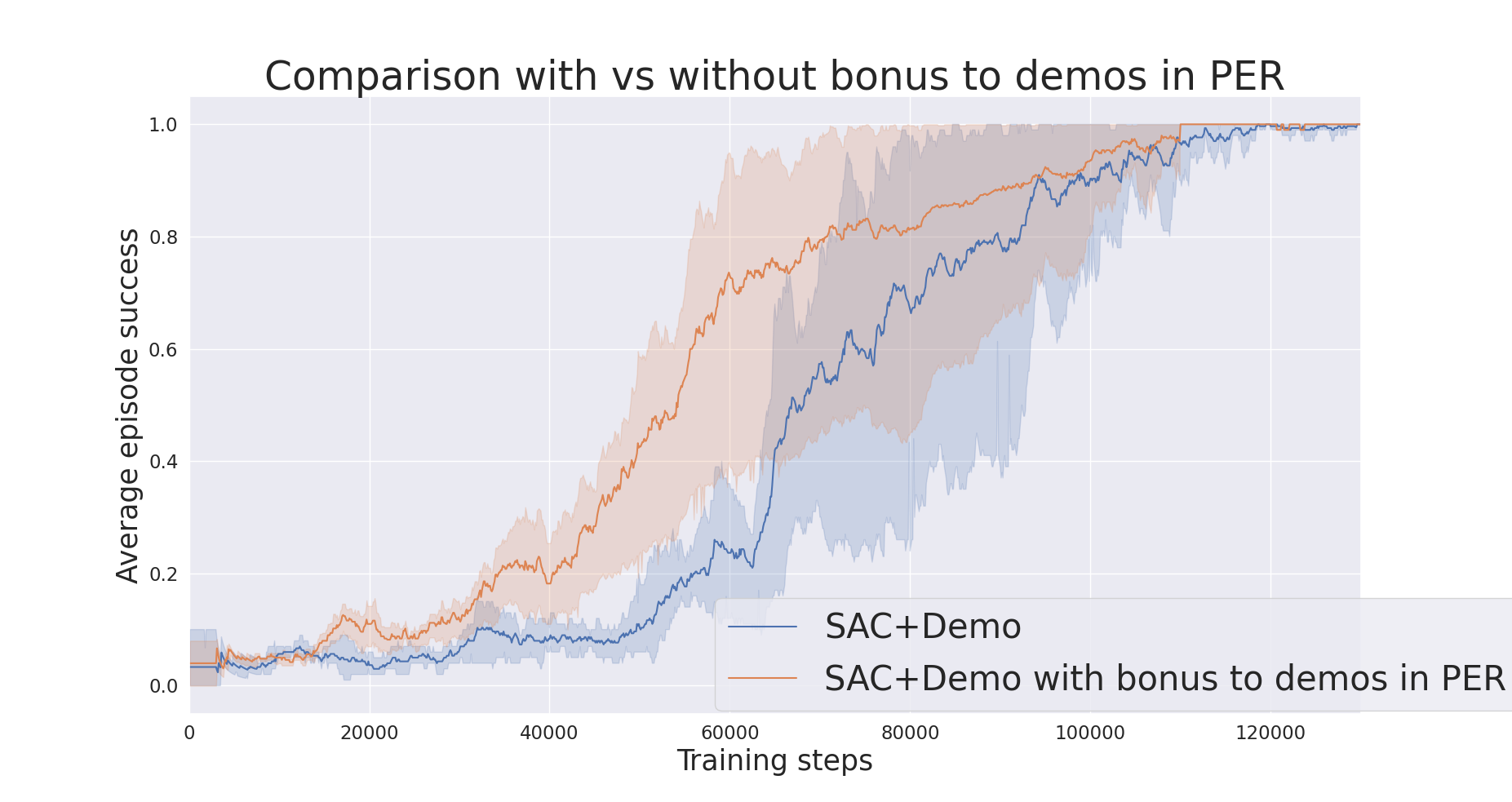}
  \end{subfigure}
  \begin{subfigure}[b]{0.495\textwidth}
    \includegraphics[width=\textwidth]{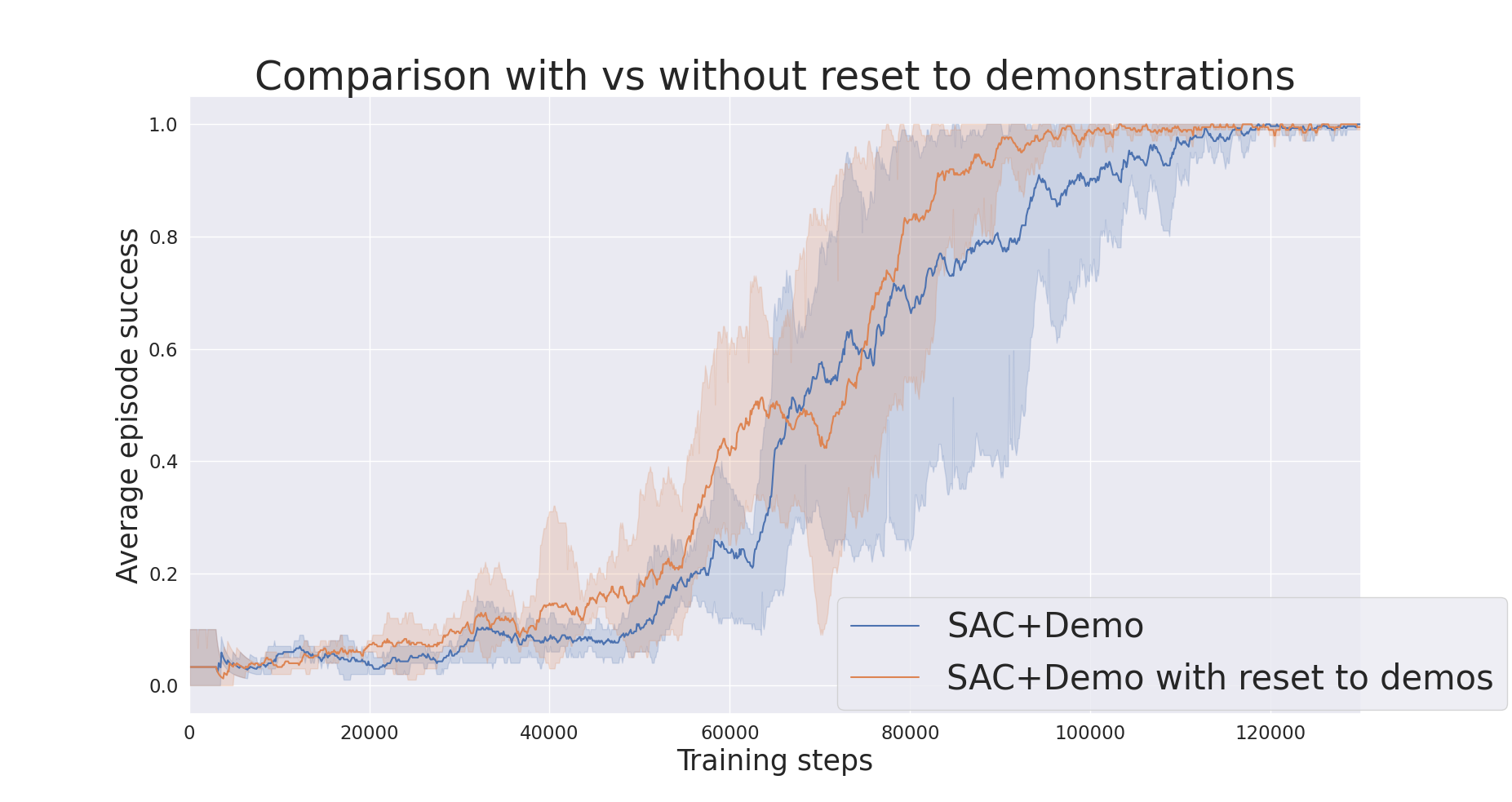}
  \end{subfigure}
  \begin{subfigure}[b]{0.495\textwidth}
    \includegraphics[width=\textwidth]{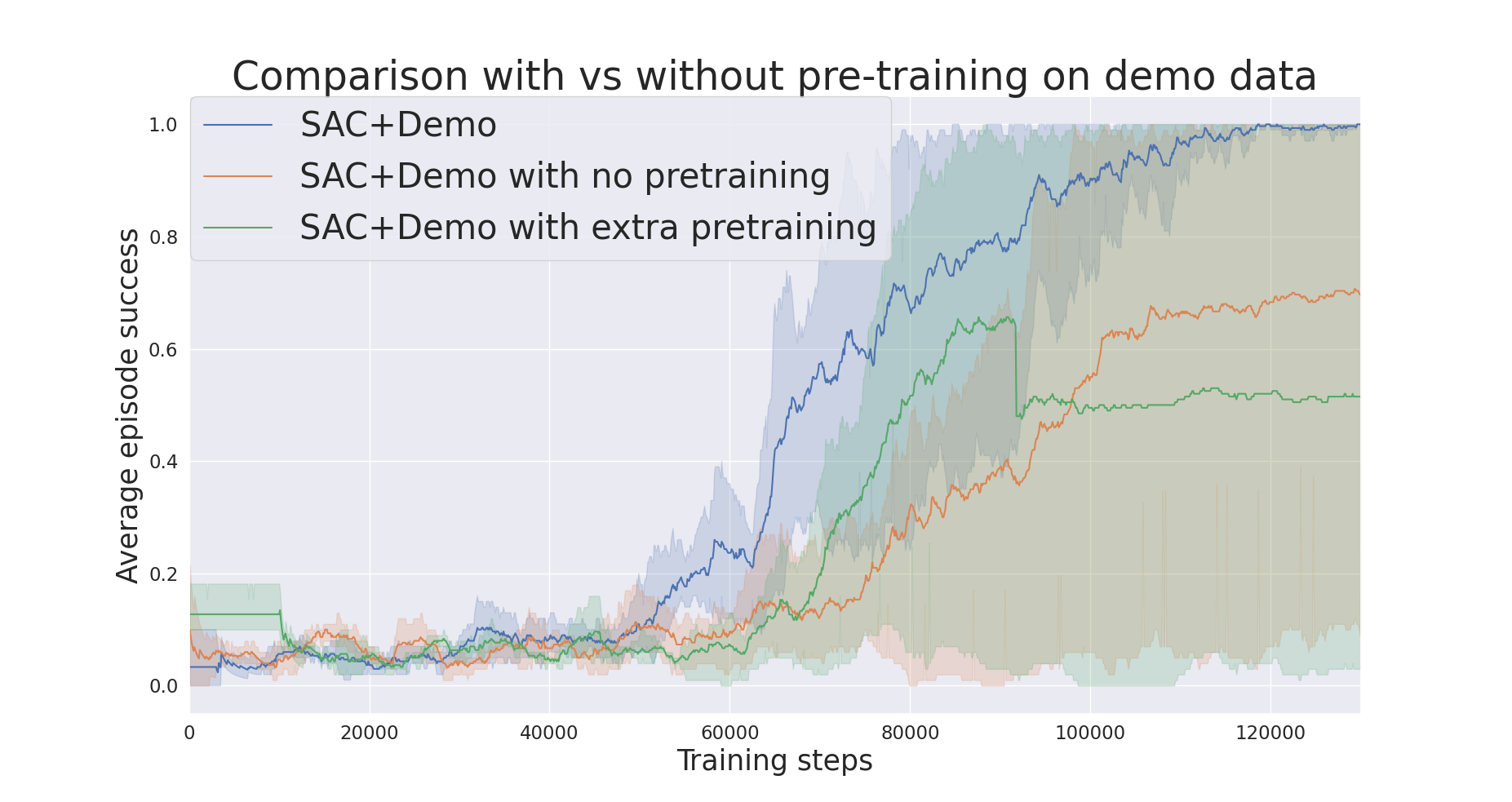}
  \end{subfigure}
  \begin{subfigure}[b]{0.495\textwidth}
    \includegraphics[width=\textwidth]{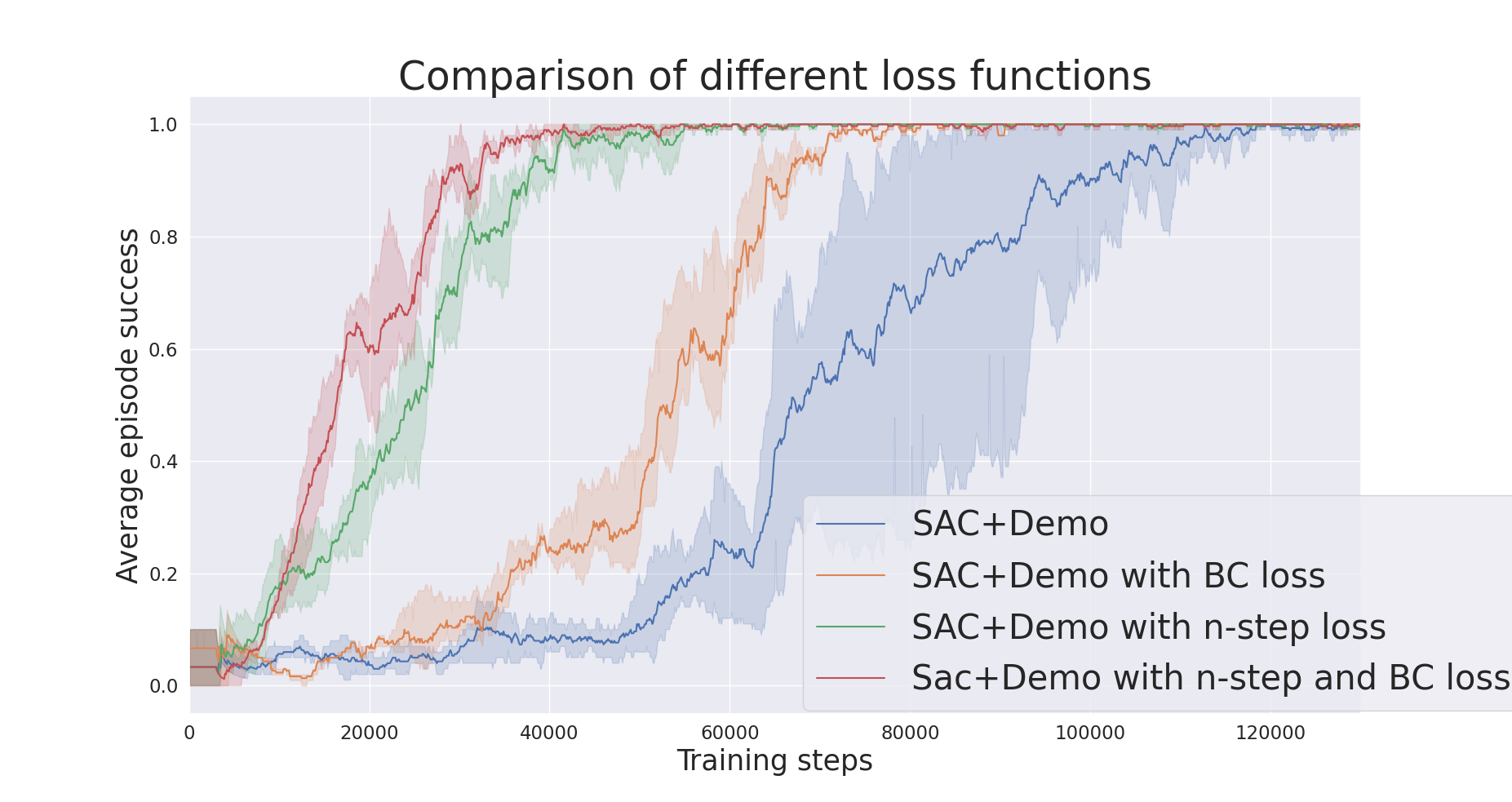}
  \end{subfigure}
  \begin{subfigure}[b]{0.495\textwidth}
    \includegraphics[width=\textwidth]{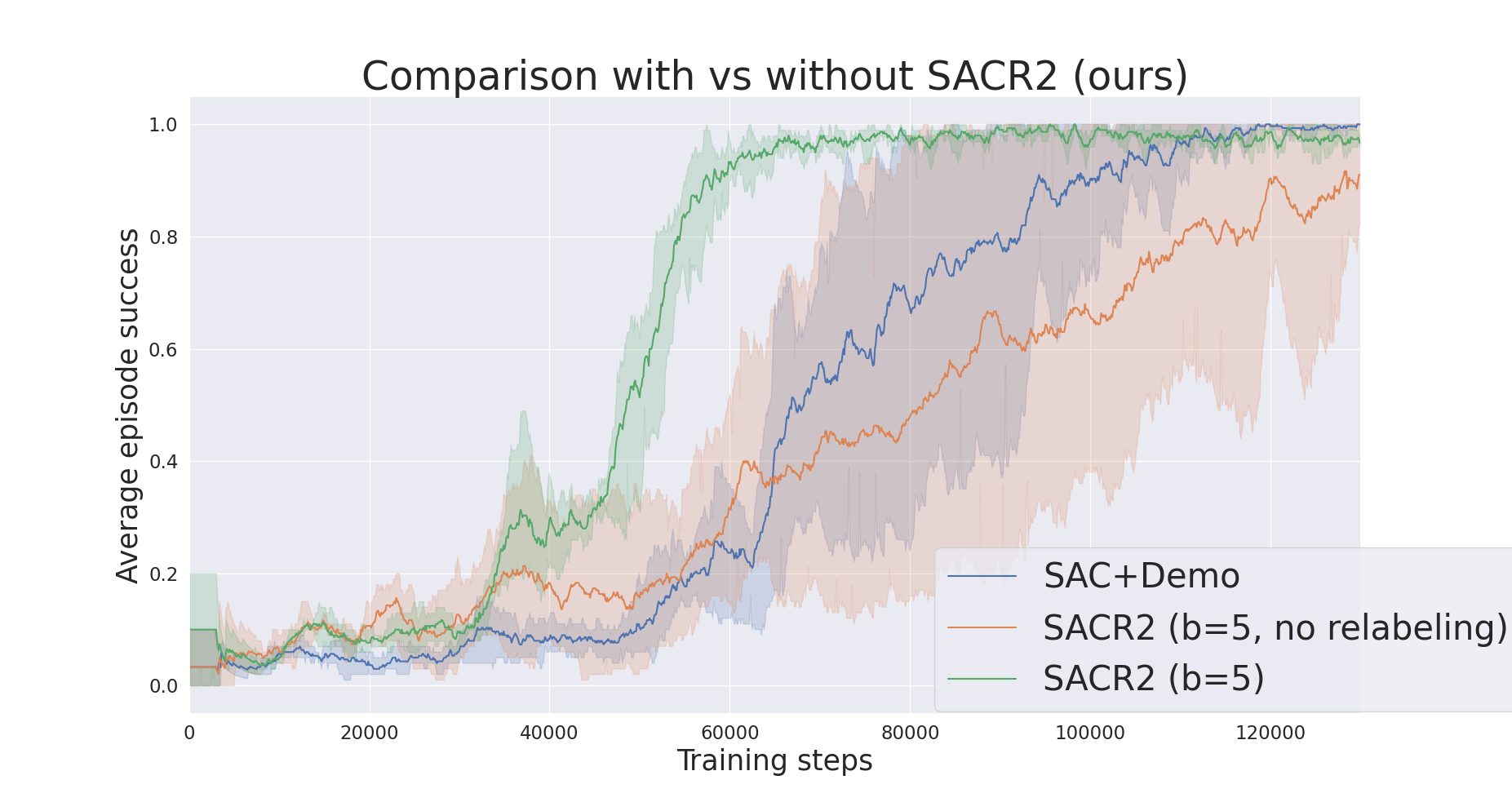}
  \end{subfigure}
  \caption{Ablation results on our baseline SAC+Demo. The results are smoothed with a rolling window of 100 episodes, and the standard error is computed on four random seeds. The n-step loss from SACfD, our approach SACR2, and the behaviour cloning loss from SACBC, are the three methods with the greatest impact on performance (in that order).}
  \label{ablation}
\end{figure}

\noindent \textbf{Loss function}. 

Following SACfD, we try a n-step return loss $\mathcal{L}_n$ to train the critic. This loss is a modified version of the loss $\mathcal{L}_1$ (see \ref{sac}) with n-step returns replacing the immediate reward. Using a larger lookahead can be particularly useful for tasks with sparse rewards, since it increases the chances of encountering a reward. We set $\lambda_n = 1$ and $n = 5$.

\begin{equation}
    \mathcal{L}_{\text{Critic}}(Q_\phi) = \mathcal{L}_1(Q_\phi) + \lambda_n \mathcal{L}_n(Q_\phi)
\end{equation}

Following SACBC, we try a behaviour cloning loss $\mathcal{L}_\text{BC}$ to train the actor.
The loss prevents the policy from deviating too much from the demonstrations, and accounts for sub-optimimality of the demonstrations by filtering out updates where the critic under-performs. However, since our demonstrations come from an expert motion planner, we decide to not include the filtering term (we ran some experiments with it and it performed significantly worse).
We set $\lambda_\text{BC} = 2$.

\begin{equation}
\label{sacbc}
\mathcal{L}_\text{BC} = \sum_i ||\pi_\theta(s_i)-a_i||_2^2 \mathds{1}_{Q_\phi(s_i,a_i)>Q_\phi(s_i,\pi_\theta(s_i))}
\end{equation}
\begin{equation}
    \mathcal{L}_{\text{Actor}}(\pi_\theta) = \mathcal{L}(\pi_\theta) + \lambda_\text{BC} \mathcal{L}_\text{BC}(\pi_\theta)
\end{equation}

The results (Figure \ref{ablation} bottom-left) show a significant increase in performance for both methods, and an even bigger increase when combined.

\textbf{One vs two buffers}.

Both SQIL and SACBC use a separate buffer to store the demonstrations, while SACfD stores them in the same buffer as the collected data. 
The main difference between these two approaches is the ratio of demonstrations in the sampled batches.
With two buffers, this ratio can be fixed (SQIL uses 50$\%$ and SACBC roughly $10\%$). With one buffer, the ratio varies from batch to batch and is on average equal to the ratio of demonstration data in the buffer if the sampling is uniform. Since we sample according to PER, the ratio is actually slightly higher, but the results (Figure \ref{ablation} top-left) show that two buffers perform better, albeit by a small margin. One practical advantage of using two buffers over one is that we can add as many demonstrations as we wish to the demonstration buffer in order to have a different replay ratio, but we introduced an equivalent amount of demonstrations to keep things fair.

\textbf{Reset to demonstrations}.

Following SACBC, we try to reset some episodes (10$\%$) to a demonstration: the position of the target ball is the same as in the demonstration, and the initial state of the robot is randomly chosen from the demonstration. This should act as a form of curriculum learning and lead to more successful episodes early on. The results (Figure \ref{ablation} middle-left) do show an initial boost in the learning process, but the improvement isn't too significant.

\textbf{Pre-training on demonstrations}.

Pre-training on demonstrations is one of the most common approaches to leverage demonstrations in the literature, and does seem like a good idea according to the results (Figure \ref{ablation} middle-right). However, too much pre-training (10000 iterations on 800 demonstrations rather than 3000 iterations on 200 demonstrations) also decreases performance. Intuitively, the agents winds up forgetting what it initially learnt from the demonstrations when it first encounters subpar trajectories from its collected experience. This is much more clear in Figure \ref{ablation2} of the appendix where we do an even more drastic pre-training (20000 iterations on 2000 demonstrations).

\textbf{Prioritized replay}.

Following SACfD, we try to modify the PER strategy with two additional terms: a term representing the actor loss, and a constant bonus applied to all transitions coming from demonstrations. 
From our limited experiments, this new strategy didn't have a great impact in terms of ratio of demonstration data in the sampled batches. With both PER and the modified PER the ratio is close to $11\%$ (we recall that $10\%$ of the transitions in the buffer come from demonstrations). However, the results (Figure \ref{ablation} top-right) do show a major initial boost during training, although the overall improvement isn't too significant.

We recall that in PER, the probability of sampling a particular transition is proportional to its priority $p_i$, which is commonly computed from the transition's temporal difference (TD) error $\delta_{i}$, for instance $p_i = \delta^{2}_{i}  + \epsilon$ where $\epsilon$ is a bonus given to all transitions.
SACfD adds a square term representing the actor loss and a second bonus $\epsilon_D$ given only to the demonstrations.
We use the same hyper-parameters as in SACfD.

\textbf{SACR2 (ours)}.

The results (Figure \ref{ablation} bottom-right) show an initial boost during training, both with and without relabeling, probably coming from the agent imitating the demonstrations more aggressively. However, without relabeling, the learning process becomes unstable, probably due to the lack of consistency on the rewards once the agent has collected enough sucessful episodes.
Overall, SACR2 increases the performance by a wide margin.

\begin{figure}[h]
\begin{center}
\includegraphics[width=\textwidth]{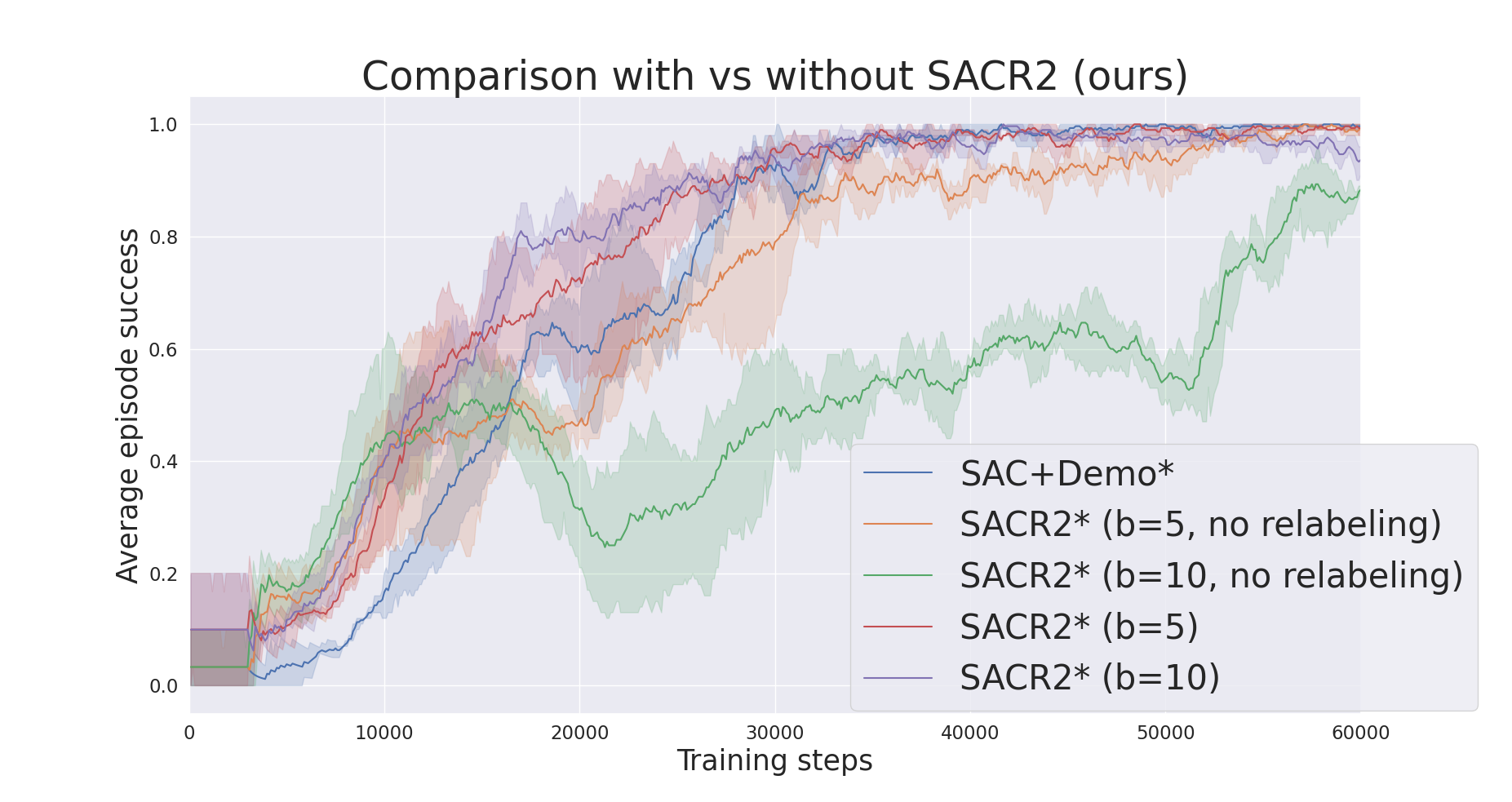}
\end{center}
\caption{Final comparison of the two more sample-efficient algorithms: SAC+Demo* and SACR2*. The results are smoothed with a rolling window of 100 episodes, and the standard error is computed on four random seeds. }
\label{sota}
\end{figure}

More experiments with different values are presented in Figure \ref{sota}. We can see that the higher $b$, the more unstable the training process without relabeling. With relabeling, there doesn't seem to be any significant difference between $b=5$ and $b=10$. We also tested $b=1$ but its impact was practically unnoticeable with respect to $b=0$.

\textbf{Best configuration}. 

Based on our ablation results, we introduce SAC+Demo* and SACR2*, which are defined as the previous algorithms plus the method that provided the greatest increase in performance: the addition of the loss functions $\mathcal{L}_\text{BC}$ and $\mathcal{L}_n$.

To answer the third question from section \ref{experimental_setup}, we carry additional experiments shown in the appendix \ref{appendix}, and the only method that furthers improves upon SAC+Demo* is SACR2*, as shown in Figure \ref{sota}. However, the improvement is very minor as SACR2* seems to be the fastest method to reach an accuracy of $90\%$, but both methods (SAC+Demo* and SACR2*) look pretty much equal afterwards.

\textbf{Learning without demonstrations}. 

\begin{figure}[h]
\begin{center}
\includegraphics[width=\textwidth]{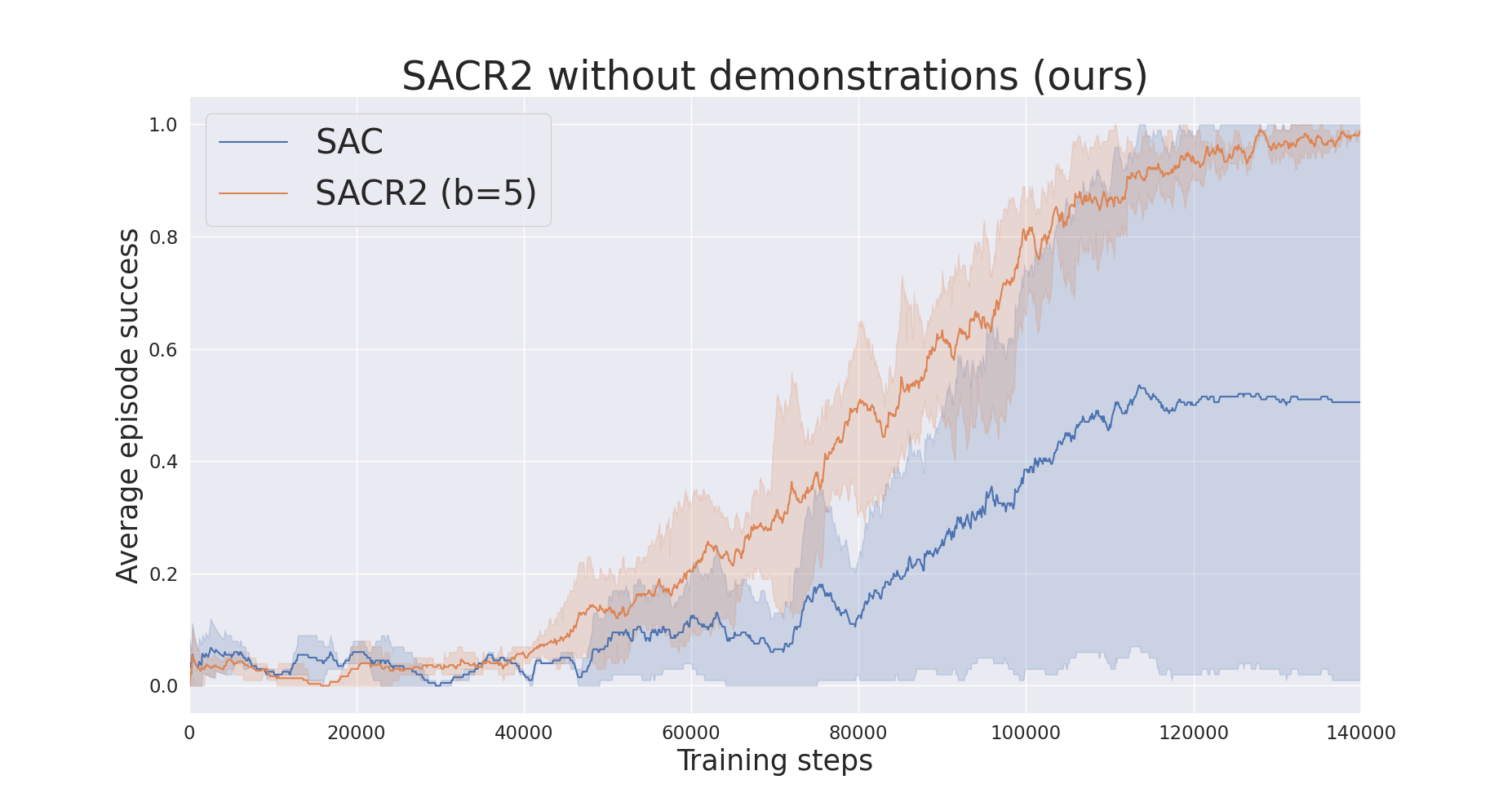}
\end{center}
\caption{Results showing that SACR2 improves upon SAC when no demonstrations are available. The results are smoothed with a rolling window of 100 episodes, and the standard error is computed on four random seeds. }
\label{nodemo}
\end{figure}

One final interesting experiment is to see whether SACR2 can also improve the performance when no demonstrations are available, by just relabeling successful episodes. Figure \ref{nodemo} shows that SACR2 actually solves the task consistently, while SAC only solves it on 2 out of the 4 runs. However, over the runs where SAC solved the task, its sample efficiency was comparable to SACR2.

\section{Discussion and Conclusion}
\label{discussion}

We propose SACR2, a generic method that can be applied to any off-policy reinforcement learning. It encourages two behaviours: imitate the expert demonstrations (if available), and imitate the past successful trajectories.
From our limited experiments, we identified three methods where the results were strong enough to advocate their use on other tasks: relabeling rewards with SACR2, the behaviour cloning loss from SACBC \citep{nair2018overcoming}, and the n-step loss from SACfD \citep{DBLP:journals/corr/VecerikHSWPPHRL17}. We show that these methods stack together, as the best results were obtained with the three methods combined.
Regarding SACR2, further analysis needs to be done on the impact of the hyper-parameters $b$ and $N$ across different tasks. Many improvements could be brought to the method, such as using a more principled value rather than a constant reward bonus, or implementing a decay or other strategy to switch the focus to the environment reward once it becomes more common.
The main limitation of our method is that it assumes that the expert demonstrations are optimal. Other methods like NAC and SACBC are able to handle sub-optimal demonstrations, which is very useful for most tasks outside of robotics, and even for more complex robotics tasks where an optimal planner is not available.
Our method also requires episodic tasks, preferably over a short horizon, since it would be difficult to relabel a continuous flow of experiences.

Our results show that SACR2 can greatly improve performance, even when no demonstrations are available and the only supervision comes from a sparse reward.
However, these results come from a single task, and we don't know how they would translate to other tasks, in particular more complex tasks where we try to increase the success rate rather than the sample efficiency. Further evaluation is also needed to clarify the impact of some of the other methods. As shown in Figures \ref{ablation} and \ref{ablation2}, some of them slightly improve upon our first baseline SAC+Demo, but actually hurt our second stronger baseline SAC+Demo*, which doesn't allow us to draw any significant conclusions.
Finally, it would be interesting to test SACR2 with another base algorithm different to SAC, for instance a Q-learning-type algorithm on a task with discrete actions.



\bibliography{iclr2022_conference}
\bibliographystyle{iclr2022_conference}

\newpage
\appendix
\section{Appendix - Additional experiments}
\label{appendix}

\begin{figure}[ht]
\centering
  \begin{subfigure}[b]{0.495\textwidth}
    \includegraphics[width=\textwidth]{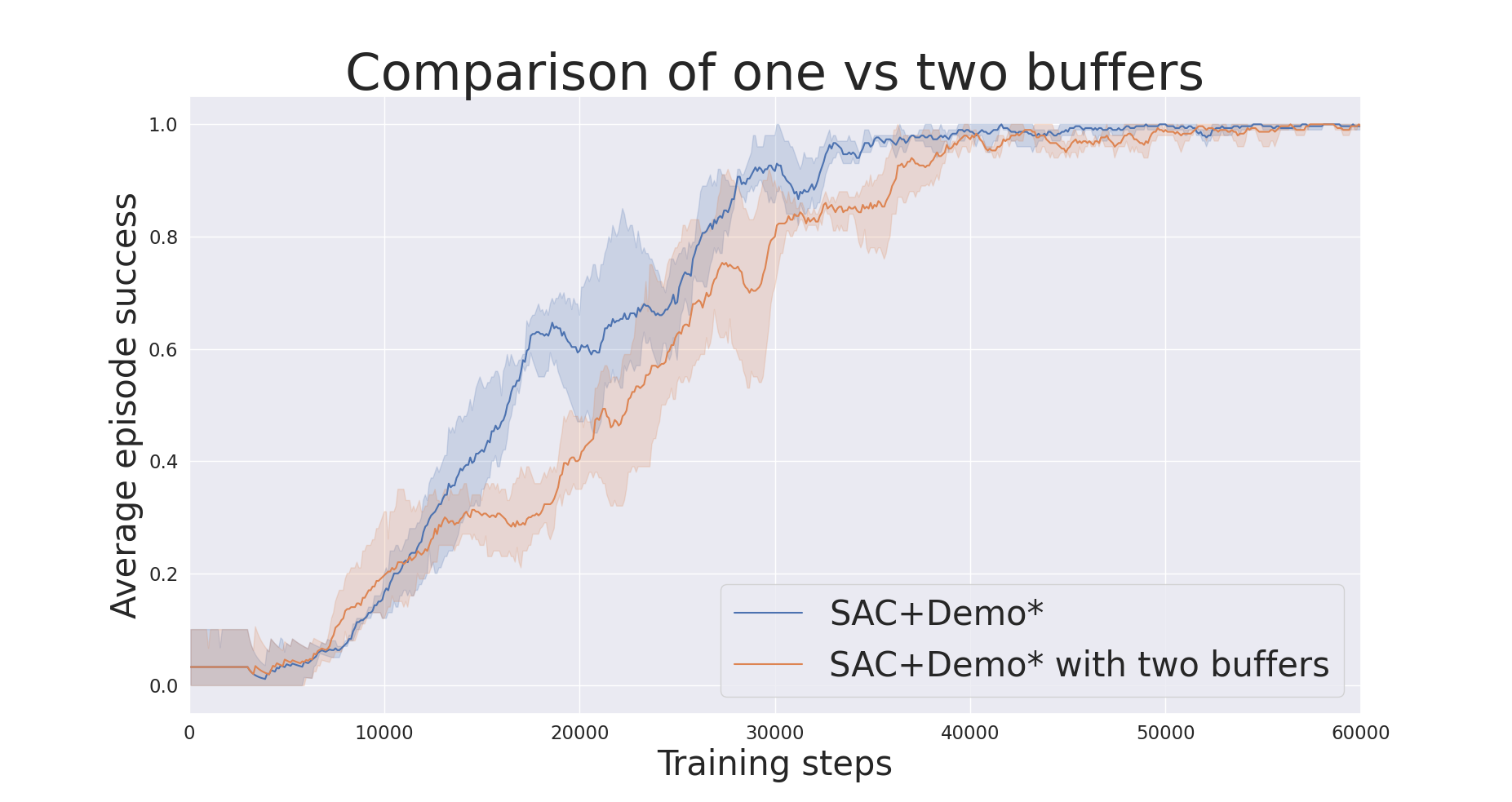}
  \end{subfigure}
  \begin{subfigure}[b]{0.495\textwidth}
    \includegraphics[width=\textwidth]{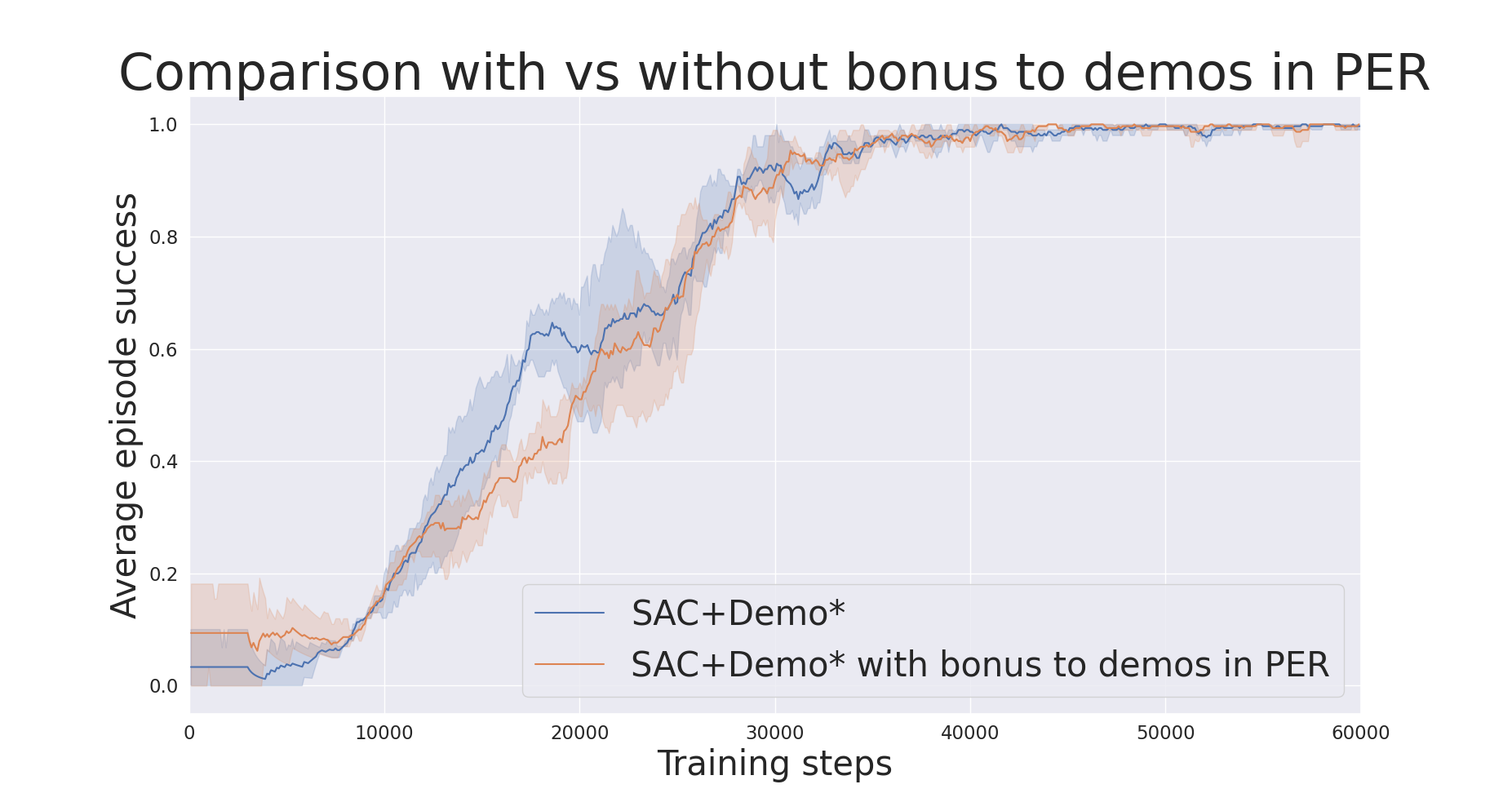}
  \end{subfigure}
  \begin{subfigure}[b]{0.495\textwidth}
    \includegraphics[width=\textwidth]{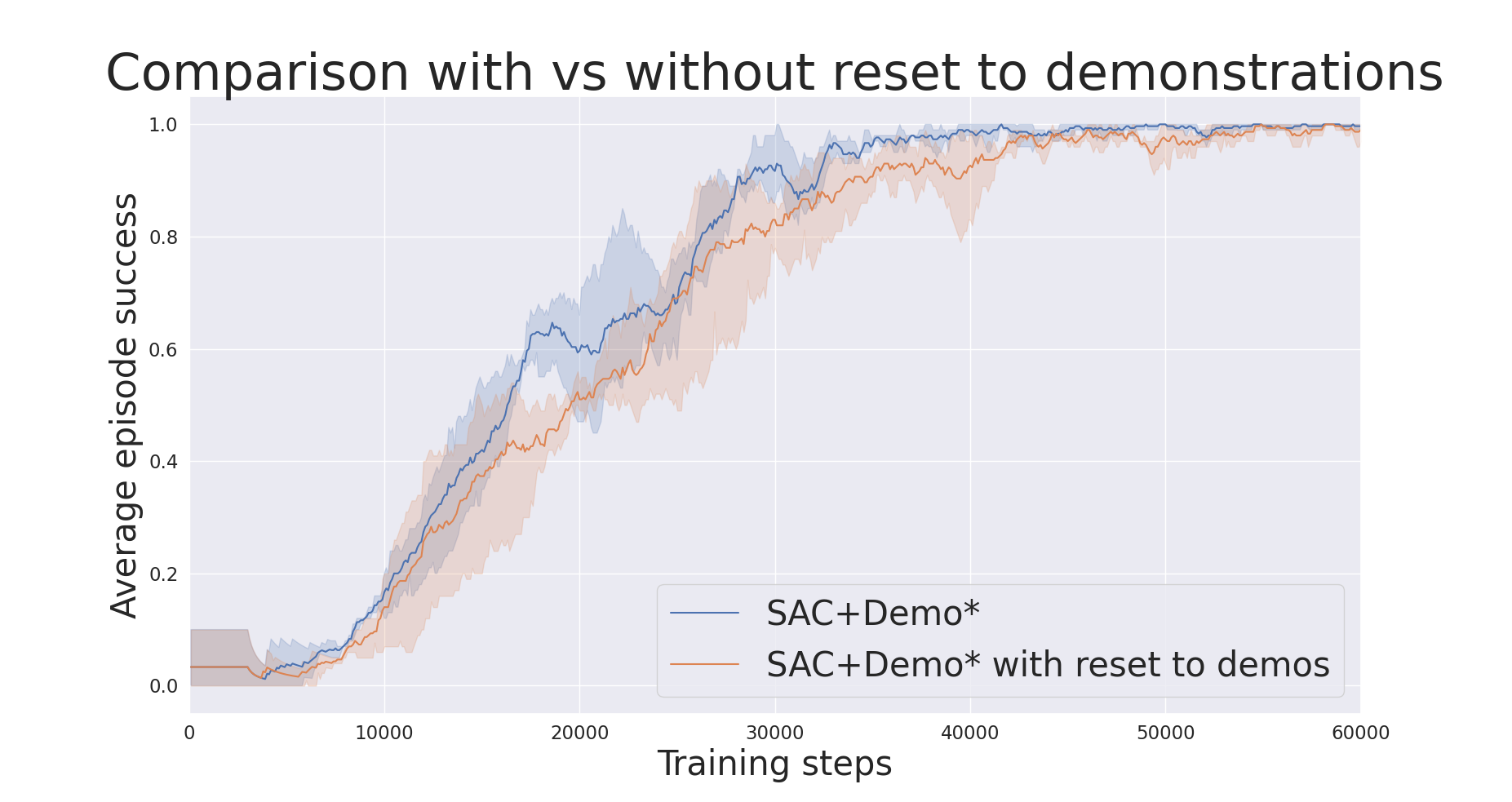}
  \end{subfigure}
  \begin{subfigure}[b]{0.495\textwidth}
    \includegraphics[width=\textwidth]{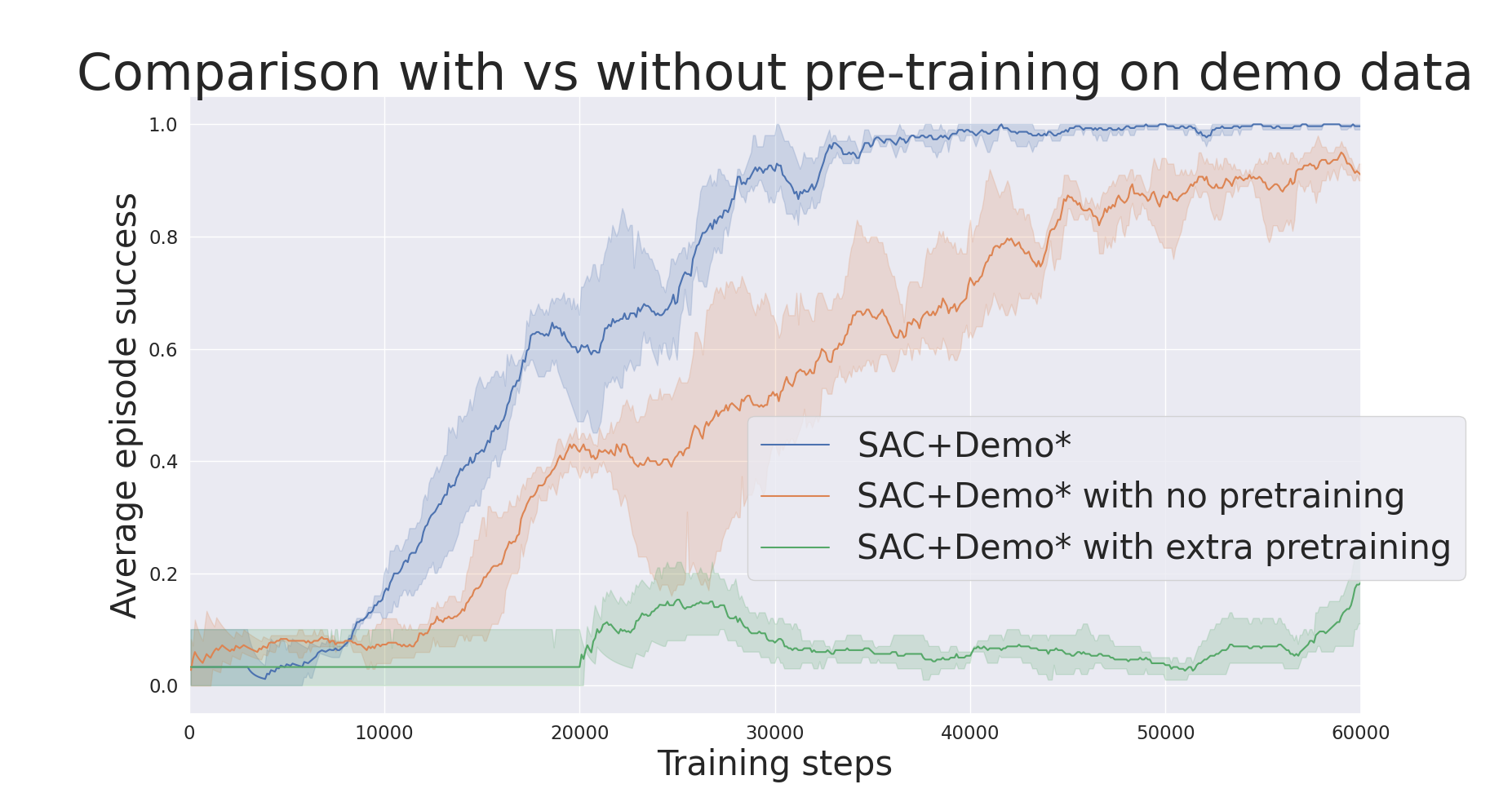}
  \end{subfigure}
  \caption{Ablation results on our improved baseline SAC+Demo*. The results are smoothed with a rolling window of 100 episodes, and the standard error is computed on four random seeds.}
  \label{ablation2}
\end{figure}

\end{document}